\documentclass[conference]{IEEEtran}
\IEEEoverridecommandlockouts
\usepackage{cite}
\usepackage{amsmath,amssymb,amsfonts}
\usepackage{algorithmic}
\usepackage{graphicx}
\usepackage{textcomp}
\usepackage{xcolor}
\def\BibTeX{{\rm B\kern-.05em{\sc i\kern-.025em b}\kern-.08em
    T\kern-.1667em\lower.7ex\hbox{E}\kern-.125emX}}

\usepackage{comment}

\usepackage{float}

\usepackage{multirow}
\usepackage{hyperref}

\usepackage{tikz}
\usetikzlibrary{positioning, arrows.meta}

\usepackage[multiple]{footmisc}
\usepackage{pifont}
\usepackage{subcaption}
\usepackage{newunicodechar}
\usepackage{tabularx}
\newunicodechar{✓}{\ding{51}}
\newunicodechar{✗}{\ding{55}}
\usepackage{xspace}
\usepackage{svg}
\usepackage[normalem]{ulem}
\usepackage{mathabx}

\usepackage[T4,T1]{fontenc}

\DeclareUnicodeCharacter{0167}{\texttstroke}
\DeclareUnicodeCharacter{0166}{\textTstroke}
\DeclareTextSymbolDefault{\texttstroke}{T4}
\DeclareTextSymbolDefault{\textTstroke}{T4}
\DeclareTextSymbol{\texttstroke}{T4}{"97}
\DeclareTextSymbol{\textTstroke}{T4}{"B7}

\usepackage{tipa}

\usepackage{pgfplots}
\pgfplotsset{compat=1.18}

\usepackage{cite}
\usepackage{amsmath,amssymb,amsfonts}
\usepackage{algorithmic}
\usepackage{graphicx}
\usepackage{textcomp}
\usepackage{xcolor}

\usepackage{multirow}
\usepackage{svg}
\usepackage{array}
\newcolumntype{?}[1]{!{\vrule width #1}}
\usepackage{hyperref}
\usepackage[multiple]{footmisc}
\usepackage{pifont}
\usepackage{subcaption}
\usepackage{newunicodechar}
\newunicodechar{✓}{\ding{51}}
\newunicodechar{✗}{\ding{55}}
\usepackage{xspace}
\usepackage{svg}
\usepackage[normalem]{ulem}
\usepackage{mathabx,booktabs}
\usepackage{array}
\newcolumntype{?}[1]{!{\vrule width #1}}
\definecolor{ashgrey}{rgb}{0.7, 0.75, 0.71}
\definecolor{trolleygrey}{rgb}{0.5, 0.5, 0.5}

\begin{document}

\title{From 2.1M Hours of Raw Broadcast Audio to Finnish Speech SSL: Confidence Filtering and Architectural Analysis at Scale}

\author{\IEEEauthorblockN{Yaroslav Getman}
\IEEEauthorblockA{\textit{Dept. lnformation and Communications Engineering} \\
\textit{Aalto University}\\
Espoo, Finland \\
yaroslav.getman@aalto.fi}
\and
\IEEEauthorblockN{Tam\'as Gr\'osz}
\IEEEauthorblockA{\textit{Walton Institute, South East Technological University, Ireland} \\
\textit{Aalto University}\\
Espoo, Finland \\
tamas.grosz@setu.ie}
\and
\IEEEauthorblockN{Tommi Lehtonen}
\IEEEauthorblockA{
\textit{Finnish
National Audiovisual Institute (KAVI)}\\
Helsinki, Finland \\
tommi.lehtonen@kavi.fi}
\and
\IEEEauthorblockN{Mikko Kurimo}
\IEEEauthorblockA{\textit{Dept. lnformation and Communications Engineering} \\
\textit{Aalto University}\\
Espoo, Finland \\
mikko.kurimo@aalto.fi}
}

\maketitle

\begin{abstract}
Broadcast archives offer scale and diversity for self-supervised learning (SSL) but contain substantial noisy, non-speech, and multilingual content. Prior work typically relies on voice activity detection or language identification (LID) to extract speech, ensuring presence rather than quality. Whether and how to filter unlabeled speech by quality for SSL, particularly from noisy broadcasts, remains underexplored. We present a large-scale Finnish SSL study starting from 2.1M hours of unconstrained broadcast audio. We propose automatic speech recognition (ASR) confidence as a proxy for pretraining data quality and show 3-7\% relative WER improvements over LID-based filtering despite a smaller training set. We further address the lack of controlled architectural comparisons in speech SSL by evaluating Transformer, Macaron-Net, Conformer, and E-Branchformer at matched parameter budgets. Our models improve upon prior Finnish SSL baselines, particularly at Base scale and out-of-domain.
\end{abstract}


\section{Introduction}
\label{sec:intro}

Self-supervised learning (SSL) has enabled effective speech representation learning, improving downstream automatic speech recognition (ASR) performance particularly in low-resource settings~\cite{baevski2020wav2vec,babu2021xlsrselfsupervisedcrosslingualspeech}. While multilingual SSL models support hundreds of languages~\cite{chen-etal-2024-towards-robust,omnilingualasrteam2025omnilingualasropensourcemultilingual}, language-specific pretraining has been shown to achieve superior performance on target language tasks~\cite{lehecka2024comparativeanalysisbilingualtrilingual,getman25_interspeech}. However, publicly available corpora are dominated by English~\cite{yt-commons,10389689}, while resources for other languages remain considerably smaller in scale and typically consist of formal or read speech~\cite{wang2021voxpopuli,pratap20_interspeech}, offering limited diversity.

To increase available pretraining data for less-resourced languages, recent work has turned to large-scale broadcast archives~\cite{malmsten2022hearing,getman25_interspeech,vaessen25_interspeech}. Such sources offer volume and variability but introduce significant challenges: limited metadata, substantial non-speech content, and multilingual segments.

Prior monolingual SSL studies have either used all available data to maximize scale or relied on selected broadcast programs and curated datasets~\cite{evain21_interspeech,PARCOLLET2024101622,lehecka22_interspeech,lp}, typically complemented by voice activity detection (VAD) segmentation~\cite{malmsten2022hearing,getman25_interspeech} or Whisper-based~\cite{10.5555/3618408.3619590} utterance extraction~\cite{vaessen25_interspeech}. 
These approaches primarily target speech extraction rather than systematic selection based on speech quality. 
While pretraining data composition has been shown to substantially affect downstream SSL performance~\cite{hsu21_interspeech}, whether and how to filter it by quality remains underexplored, especially for noisy broadcast archives.

Confidence-based data selection is well established in semi-supervised and self-training pipelines, with model scores used to filter pseudo-labels~\cite{kahn2020selftraining,park20d_interspeech}. \cite{10447489} applied diarization confidence to filter data for self-supervised speaker representation learning. 
However, to our knowledge, ASR confidence has not been investigated for SSL pretraining data selection.

While Transformer~\cite{10.5555/3295222.3295349} encoders are widely used in SSL frameworks~\cite{baevski2020wav2vec,10.1109/TASLP.2021.3122291,9814838,pmlr-v162-baevski22a}, alternatives such as Conformer~\cite{gulati20_interspeech} and E-Branchformer~\cite{10022656} have been adopted in self-supervised settings~\cite{9688253,pmlr-v162-chiu22a,chen-etal-2024-towards-robust}. Controlled comparisons across encoder architectures exist for supervised ASR~\cite{peng23b_interspeech}, yet prior SSL studies have varied in both architecture and model size. For instance, Conformer-based and Transformer-based wav2vec 2.0~\cite{baevski2020wav2vec} variants in \cite{wang-etal-2020-fairseq} used 618M and 317M parameters respectively, making it unclear whether gains stem from architecture or capacity.

In this work, we present a large-scale study of speech SSL for Finnish, addressing both data filtering and encoder architecture. 
Starting from 2.1M hours of raw, unconstrained broadcast material, we apply a multi-stage pipeline combining VAD, deduplication, and ASR confidence-based filtering, yielding 471K hours of pretraining data.
To our knowledge, this represents the largest academic effort in monolingual SSL pretraining to date.
We make the following contributions:
\begin{enumerate}
    \item We propose and systematically evaluate ASR confidence-based filtering for SSL pretraining data selection from unconstrained broadcast archives, showing 3-7\% relative WER improvements over language-identification-based filtering despite a smaller retained training set.

    
    \item We conduct, to our knowledge, the first parameter-controlled comparison of Transformer, Macaron-Net~\cite{lu2019understanding}, Conformer, and E-Branchformer encoders within a unified SSL framework, scaling from 95M to 965M parameters. Convolution-augmented encoders yield up to 37\% relative WER reduction (WERR) over Transformer at Base scale, where architectural choice has the greatest impact.

    \item While the broadcast archives cannot be redistributed due to copyright and broadcast licensing restrictions, we release our models publicly\footnote{\resizebox{.92\columnwidth}{!}{\url{https://github.com/aalto-speech/large-scale-monolingual-speech-foundation-models}}}, which improve upon previous Finnish SSL baselines, particularly in parameter-constrained and out-of-domain scenarios.
    
\end{enumerate}









\begin{table*}[ht!]
\centering
\caption{
Comparison of encoder architectures. Equations illustrate the sequence of sub-modules and residual connections within a single encoder block. $\mathrm{Merge}(\cdot)$ denotes the depth-wise convolutional fusion module. Layer normalizations are omitted for clarity.
}
\begin{tabularx}{\textwidth}{@{} p{0.20\textwidth} p{0.20\textwidth} p{0.21\textwidth} X @{}}
\toprule
\textbf{Transformer}~\cite{10.5555/3295222.3295349} &
\textbf{Macaron-Net}~\cite{lu2019understanding} &
\textbf{Conformer}~\cite{gulati20_interspeech} &
\textbf{E-Branchformer}~\cite{10022656} \\
\midrule
$\begin{aligned}
x'_i &= x_i + \mathrm{MHSA}(x_i) \\
y_i &= x'_i + \mathrm{FFN}(x'_i)
\end{aligned}$ &
$\begin{aligned}
\tilde{x}_i &= x_i + \tfrac{1}{2}\mathrm{FFN}(x_i) \\
x'_i &= \tilde{x}_i + \mathrm{MHSA}(\tilde{x}_i) \\
y_i &= x'_i + \tfrac{1}{2}\mathrm{FFN}(x'_i)
\end{aligned}$ &
$\begin{aligned}
\tilde{x}_i &= x_i + \tfrac{1}{2}\mathrm{FFN}(x_i) \\
x'_i &= \tilde{x}_i + \mathrm{MHSA}(\tilde{x}_i) \\
x''_i &= x'_i + \mathrm{Conv}(x'_i) \\
y_i &= x''_i + \tfrac{1}{2}\mathrm{FFN}(x''_i)
\end{aligned}$ &
$\begin{aligned}
\tilde{x}_i &= x_i + \tfrac{1}{2}\mathrm{FFN}(x_i) \\
x'_{ig} &= \mathrm{MHSA}(\tilde{x}_i),\quad x'_{il} = \mathrm{cgMLP}(\tilde{x}_i) \\
x''_i &= \tilde{x}_i + \mathrm{Merge}(x'_{ig}, x'_{il}) \\
y_i &= x''_i + \tfrac{1}{2}\mathrm{FFN}(x''_i)
\end{aligned}$ \\
\bottomrule
\end{tabularx}
\label{tab:w2v2_backends}
\end{table*}

\section{Data}
\label{sec:data}

We use the Radio and Television Archive (RTVA) of the Finnish National Audiovisual Institute (KAVI), comprising broadcasts from 26 TV channels, 85 radio stations, and online media content, totaling 2.1M hours. The recordings span 2009-2025, with some sources recorded continuously and others sampled periodically. 
Content is highly variable, ranging from formal news broadcasts to entertainment programs, advertisements, and music, and metadata is limited. 


We also include the training sets of three publicly available Finnish speech datasets: the Finnish subset of VoxPopuli~\cite{wang2021voxpopuli} (14.3K hours), the Finnish Parliament~\cite{finparl} (FP, 3.1K hours, fully transcribed), and Lahjoita puhetta~\cite{Moisio2022LahjoitaPA} (LP, 3.2K hours, partially transcribed). These datasets are processed using the same filtering pipeline as the archival data for consistency. The transcribed portions of FP and LP (4.6K hours in total) are also used for supervised fine-tuning. For evaluation on real-world broadcast conditions, we created a manually transcribed test set from KAVI, comprising two one-hour radio broadcasts from stations Nova and Basso.

\section{Methodology}
\label{sec:methodology}

\subsection{Self-supervised pretraining framework}

We adopt the wav2vec 2.0 framework~\cite{baevski2020wav2vec}, which remains widely used for non-English monolingual SSL~\cite{evain21_interspeech,malmsten2022hearing,lehecka22_interspeech,PARCOLLET2024101622,10096318,lehecka24_interspeech,lp,getman25_interspeech,vaessen25_interspeech}, enabling direct comparison with prior work. 
We opt for three model scales following~\cite{babu2021xlsrselfsupervisedcrosslingualspeech}: Base (95M parameters), Large (317M parameters), and X-Large (965M parameters). Within each scale, we explore alternative encoder backends at matched parameter counts to ensure performance gains stem from architectural design rather than raw capacity. We additionally fix the pretraining data across backends to isolate architectural effects and evaluate across six test sets spanning in-domain and out-of-domain conditions to ensure conclusions are not specific to a single domain.

\subsection{Encoder architectures}

We explore four encoder backends for wav2vec 2.0, summarized in Table~\ref{tab:w2v2_backends}. The standard wav2vec 2.0 uses the Transformer~\cite{10.5555/3295222.3295349}, where each layer consists of a multi-head self-attention (MHSA) module followed by a feed-forward network (FFN), each with residual connections and layer normalization.

Macaron-Net~\cite{lu2019understanding} modifies the Transformer by replacing the single FFN with two half-step FFNs before and after MHSA, with inner dimensionality halved to maintain parameter parity. This sandwich structure is incorporated in architectures such as Conformer~\cite{gulati20_interspeech} and E-Branchformer~\cite{10022656}, yet we evaluate Macaron-Net as a standalone encoder to isolate the effect of the dual-FFN configuration from other architectural changes.

Conformer~\cite{gulati20_interspeech} augments the Macaron structure by inserting a convolution module between self-attention and the final FFN. This sequential design captures both global dependencies via attention and local patterns through convolution.

In contrast, E-Branchformer~\cite{10022656} processes global (MHSA) and local (convolutional gated MLP) features in parallel branches. These representations are merged via a depthwise convolutional fusion module, allowing the encoder to extract complementary acoustic information independently.

\subsection{Filtering pipeline}

We apply a multi-stage filtering pipeline to address challenges inherent to broadcast data.
First, we apply voice activity detection (VAD) using pyannote.audio~\cite{bredin23_interspeech}, yielding 1.2M hours of speech segments, split into maximum 30-second chunks.

Since broadcasts contain substantial repetitive content, including jingles, advertisements, and rebroadcast programs, VAD alone is insufficient. We next perform transcript-based deduplication using ASR transcripts generated by the Finnish wav2vec 2.0 Large model~\cite{getman25_interspeech}, retaining a single instance per unique transcript, as identical transcripts reliably capture rebroadcast content regardless of minor acoustic variations. Analysis of the ASR transcripts revealed that some phrases repeat tens of thousands of times across channels and time periods, confirming the necessity of this step. While this may reduce acoustic diversity for repeated phrases spoken by different speakers, we prioritize removing extreme redundancy.
This reduces the total data from 1.2M to 890K hours.

Beyond deduplication, we use ASR confidence as a proxy for pretraining data quality. The intuition is that a well-trained ASR model assigns higher CTC log-probabilities to acoustically clear speech rather than to noisy or irrelevant audio. We define the confidence metric $C_{\text{char}}$ as the mean log-probability of the predicted frame-wise character sequence, excluding blank tokens ($\epsilon$), as blank exclusion with mean aggregation has been shown to yield more robust confidence estimates~\cite{gitman23_interspeech}:


\begin{equation}
        C_{\text{char}} = \frac{1}{|S|}\sum_{t\in S}\log P(\hat{y}_t \mid X), \quad S = \{t \mid \hat{y}_t \neq \epsilon\}
\end{equation}

where $\hat{y}_t = \arg \max_{y} P(y \mid X)$ denotes the predicted label at frame $t$ without collapsing repeats. A sample is retained if $C_{\text{char}} \geq \tau$ for threshold $\tau$ (selected in Section~\ref{subsec:filtering}).

For comparison, we also implement language-identification-based (LID) filtering using an ECAPA-TDNN model\footnote{https://huggingface.co/speechbrain/lang-id-voxlingua107-ecapa} for audio-based LID and fastText~\cite{joulin2016bag,joulin2016fasttext} on ASR transcripts for text-based LID, with segments retained if identified as Finnish by both models. Analysis of the LID results confirmed substantial multilingual content in the archives, with English as the second most common language ($\sim$80K hours).


\section{Experiments}
\label{sec:experiments}

\subsection{Experimental setup}

We pretrain models on 512 AMD MI250X GPUs and fine-tune them on 128 GPUs using Fairseq~\cite{ott2019fairseq}. Since Macaron-Net and E-Branchformer are not natively supported in Fairseq, we implement them as extensions following their original architectural specifications. 

Base models are pretrained with a learning rate of 6e-4 and a dynamic batch size of 3 minutes per GPU for 125K steps. Large models use learning rate 4e-4 and batch size 2 minutes per GPU for 167K steps. X-Large models use learning rate 1e-4 and batch size of 1 minute per GPU for 400K steps. 

Models are fine-tuned for ASR using CTC loss~\cite{ctc}, with the CNN feature encoder frozen. We select learning rates from 1e-5 to 1e-4 based on validation performance. Base models use a batch size of 2.5 minutes per GPU and 16K updates. Large models are fine-tuned with a batch size of 2 minutes per GPU for 25K steps. X-Large models use approximately 1 minute per GPU and train for 40-45K steps (exact batch size and step count vary by architecture due to memory constraints).

We evaluate ASR performance using word error rate (WER). For in-domain evaluation, we report results on the test splits of the two fine-tuning corpora, Finnish Parliament (FP)~\cite{finparl}, which includes two distinct test sets, and Lahjoita puhetta (LP)~\cite{Moisio2022LahjoitaPA}. To assess generalization beyond the fine-tuning domains, we evaluate on Common Voice 16.1~\cite{ardila-etal-2020-common} and FLEURS~\cite{fleurs} benchmarks. Finally, we evaluate performance under challenging real-world broadcast conditions using the KAVI test set.

\subsection{Data filtering experiments}
\label{subsec:filtering}

Given the unconstrained nature of the broadcast archives, selecting an effective filtering pipeline is critical for successful SSL pretraining. To analyze the impact of filtering choices, we train wav2vec 2.0 Base Transformer models on subsets of the deduplicated VAD data with: (i) no further filtering; (ii) LID-based filtering; and (iii) ASR confidence-based filtering with thresholds $\tau \in \{-0.3, -0.25, -0.2, -0.15, -0.1\}$. Models are fine-tuned on the 4.6K-hour supervised set and evaluated on the validation splits of FP, LP, Common Voice, and FLEURS.

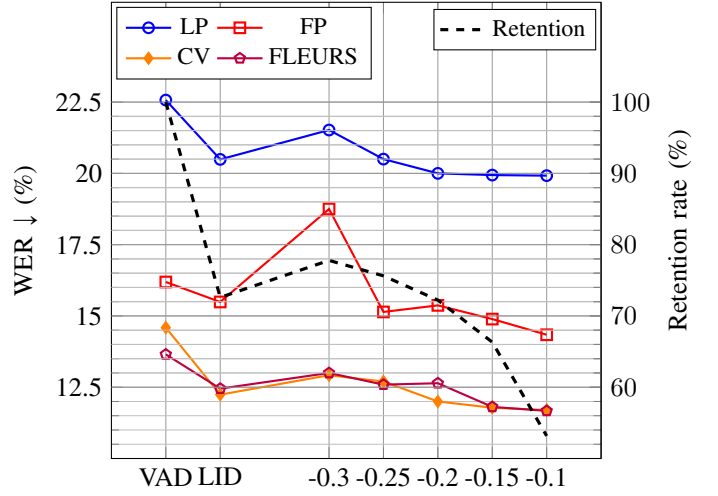
\begin{figure}[t]
\centering
\begin{tikzpicture}
\begin{axis}[
    name=plot1,
    width=0.91\linewidth,
    height=0.42\textwidth,
    ylabel={WER $\downarrow$ (\%)},
    grid=major,
    xmin=-0.50, xmax=-0.05,
    ymin=10, ymax=26,
    ytick={10,12.5,15,17.5,20,22.5},
    yticklabels={,12.5,15,17.5,20,22.5},
    xtick={-0.45, -0.40, -0.30, -0.25, -0.20, -0.15, -0.10},
    xticklabels={VAD, LID, -0.3, -0.25, -0.2, -0.15, -0.1},
    grid=both,
    minor tick num=4,
    grid style={line width=0.3pt, draw=gray!80},
    minor grid style={line width=0.25pt, draw=gray!50},
    mark size=2.0pt,
    legend style={font=\small, 
                  at={(0.01, 0.99)},
                  anchor=north west,
                  legend columns=2,
                  },
]

\addplot[mark=o, blue, thick] coordinates {
    (-0.45,22.57) (-0.40,20.49) (-0.30,21.52) (-0.25,20.50)
    (-0.20,20.00) (-0.15,19.94) (-0.10,19.92)};
\addlegendentry{LP}

\addplot[mark=square, red, thick] coordinates {
    (-0.45,16.19) (-0.40,15.49) (-0.30,18.75) (-0.25,15.14)
    (-0.20,15.37) (-0.15,14.89) (-0.10,14.34)};
\addlegendentry{FP}

\addplot[mark=diamond*, orange, thick] coordinates {
    (-0.45,14.59) (-0.40,12.24) (-0.30,12.92) (-0.25,12.70)
    (-0.20,12.00) (-0.15,11.78) (-0.10,11.70)};
\addlegendentry{CV}

\addplot[mark=pentagon, purple, thick] coordinates {
    (-0.45,13.65) (-0.40,12.45) (-0.30,13.00) (-0.25,12.59)
    (-0.20,12.64) (-0.15,11.81) (-0.10,11.67)};
\addlegendentry{FLEURS}

\end{axis}

\begin{axis}[
    width=0.91\linewidth,
    height=0.42\textwidth,
    xlabel={},
    ylabel={Retention rate (\%)},
    ylabel style={black},
    axis y line*=right,
    axis x line=none,
    xmin=-0.50, xmax=-0.05,
    ymin=50, ymax=114,
    ytick={60,70,80,90,100},
    xtick=\empty,
    grid=both,
    minor tick num=4,
    grid style={line width=0.3pt, draw=gray!80},
    minor grid style={line width=0.25pt, draw=gray!50},
    legend style={font=\small, 
                  at={(0.99, 0.94)},
                  anchor=east,
                  legend columns=1,
                  inner sep=1.8pt,
                  },
]
\addplot[
    black, dashed, very thick,
    mark=None, mark size=3pt, mark options={fill=white, draw=black}
] coordinates {
    (-0.45, 100)
    (-0.40, 72.60)
    (-0.30, 77.79)
    (-0.25, 75.62)
    (-0.20, 72.22)
    (-0.15, 66.32)
    (-0.10, 53.17)
};
\addlegendentry{Retention}
\end{axis}
\end{tikzpicture}
\caption{
Impact of filtering criteria on downstream validation WER and pretraining data retention rate. All confidence thresholds ($C_{\text{char}} \geq \tau$) are applied to deduplicated VAD segments.
}
\label{fig:data_filtering}
\end{figure}

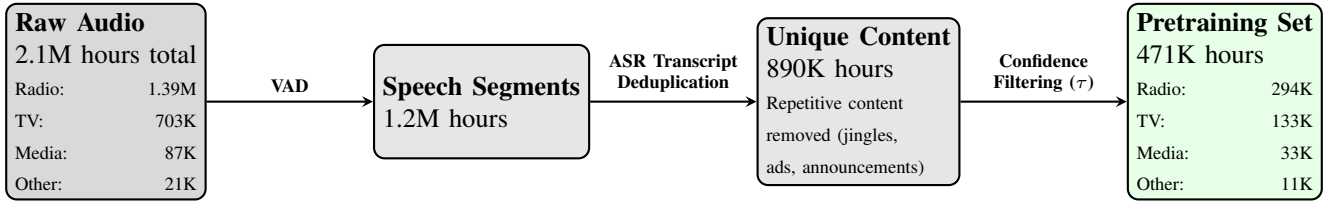
\begin{figure*}[h]
\centering
\begin{tikzpicture}[
    node distance=2.2cm,
    start/.style={draw, thick, fill=black!15, rounded corners, minimum width=2.2cm, minimum height=1.5cm, align=left},
    intermediate/.style={draw, thick, fill=black!10, rounded corners, minimum width=2.5cm, minimum height=1.5cm, align=left},
    final/.style={draw, thick, fill=green!10, rounded corners, minimum width=2.5cm, minimum height=1.5cm, align=left},
    arrow/.style={-stealth, thick, black}
]

\node (raw) [start] {
    \textbf{Raw Audio}\\
    2.1M hours total\\
    \scriptsize Radio:\hfill1.39M\\
    \scriptsize TV:\hfill703K\\
    \scriptsize Media:\hfill87K\\
    \scriptsize Other:\hfill21K
};

\node (vad) [intermediate, right=of raw] {
    \textbf{Speech Segments}\\
    1.2M hours
};

\node (deduplication) [intermediate, right=of vad] {
    \textbf{Unique Content}\\
    890K hours\\
    \scriptsize Repetitive content\\
    \scriptsize removed (jingles,\\
    \scriptsize ads, announcements)
};

\node (final) [final, right=of deduplication] {
    \textbf{Pretraining Set}\\
    471K hours\\
    \scriptsize Radio:\hfill294K\\
    \scriptsize TV:\hfill133K\\
    \scriptsize Media:\hfill33K\\
    \scriptsize Other:\hfill11K
};

\path (raw) edge [arrow] node[above, font=\scriptsize\bfseries, align=center] {VAD} (vad);
\path (vad) edge [arrow] node[above, font=\scriptsize\bfseries, align=center] {ASR Transcript\\Deduplication} (deduplication);
\path (deduplication) edge [arrow] node[above, font=\scriptsize\bfseries, align=center] {Confidence\\Filtering ($\tau$)} (final);

\end{tikzpicture}
\caption{Pretraining data preparation pipeline. We reduce 2.1M hours of raw broadcast archives to a high-quality 471K-hour subset through VAD, transcript-based deduplication, and the confidence-based filtering threshold $\tau=-0.1$ optimized in Figure~\ref{fig:data_filtering}.
}
\label{fig:pipeline}
\end{figure*}

Figure~\ref{fig:data_filtering} shows the impact of filtering on downstream ASR performance and data retention. At similar retention rates (\textasciitilde72\%), confidence filtering with $\tau = -0.2$ achieves lower WER than LID-only filtering on 3 of 4 benchmarks. 
Performance generally stabilizes in the range $\tau \in [-0.15, -0.1]$, with $\tau = -0.1$ consistently yielding the lowest WER, particularly on FP. We therefore select $\tau = -0.1$ for subsequent experiments as the strictest threshold that avoids performance degradation while still retaining sufficient data volume (471K hours, or 53\% of deduplicated data). 
Notably, confidence-based filtering naturally removes both low-quality target language segments and most non-target language content: LID analysis of the final 471K-hour set shows $\sim$99\% Finnish content, eliminating the need for explicit language filtering. The final data preparation pipeline is summarized in Figure~\ref{fig:pipeline}.

\begin{table}[t]
\centering
\caption{WER (\%) for wav2vec 2.0 Base variants across Finnish benchmarks. The best results within the same encoder backend are \underline{underlined}, and the best overall results are shown \textbf{in bold}.}
\resizebox{\columnwidth}{!}{%
\begin{tabular}{@{}lcccc | ccc | ccc@{}}
\toprule
\multirow{2}{*}{\textbf{Backend}} & \multirow{2}{*}{\textbf{L}} & \multirow{2}{*}{\textbf{Enc.}} & \multirow{2}{*}{\textbf{FFN}} & \multirow{2}{*}{\textbf{Params}} & \multicolumn{3}{c|}{\textbf{In-domain}} & \multicolumn{3}{c}{\textbf{Out-of-domain}} \\
\cmidrule(lr){6-8} \cmidrule(lr){9-11}
 & & & & & \textbf{LP} & \textbf{FP'16} & \textbf{FP'20} & \textbf{CV} & \textbf{FLEURS} & \textbf{KAVI} \\
\midrule
Transformer & 6 & 1024 & 4096 & 90.6M & 28.30 & 14.08 & 11.58 & 15.60 & 17.13 & 31.49 \\
            & 12 & 768 & 3072  & 95.0M & 22.28 & 10.72 & 8.70  & 9.74  & 13.02 & 24.85 \\
            & 18 & 768 & 1536  & 95.1M & \underline{21.33} & \underline{10.64} & \underline{8.66}  & 8.99  & \underline{12.71} & \underline{23.91} \\
            & 24 & 576 & 2048  & 96.4M & 21.66 & 10.69 & 8.73  & \underline{8.77}  & 12.85 & 24.57 \\
\midrule
Macaron-Net & 12 & 768 & 1536  & 95.1M & 21.80 & 10.58 & 9.10  & 8.62  & 12.54 & 25.22 \\
            & 22 & 576 & 1152  & 95.6M & \underline{19.41} & \underline{9.62} & \underline{7.92}  & \underline{7.12}  & \underline{11.07} & \underline{23.51} \\
\midrule
Conformer   & 12 & 576 & 2048  & 96.7M & 18.89 & 10.08 & 8.07  & 6.46  & 11.20 & 23.71 \\
            & 17 & 512 & 1536  & 96.5M & \underline{\textbf{18.50}} & \underline{9.83} & \underline{7.94}  & \underline{\textbf{5.69}}  & \underline{\textbf{10.56}} & \underline{22.71} \\
\midrule
E-Branch.   & 12 & 576 & 1280  & 98.4M & 19.14 & 10.10 & \underline{\textbf{7.80}}  & 6.91  & 11.53 & 22.89 \\
            & 17 & 512 & 1024  & 96.6M & \underline{18.80} & \underline{\textbf{9.54}}  & 7.85  & \underline{6.19}  & \underline{11.09} & \underline{\textbf{22.44}} \\
\bottomrule
\end{tabular}
}
\label{tab:architecture_exp_res}
\end{table}


\begin{table*}[t]
\centering
\caption{WER (\%) across scales and encoder backends. Best results per scale are \underline{underlined} and best overall \textbf{in bold}. LP = Lahjoita puhetta (colloquial Finnish), FP = Finnish Parliament, CV = Common Voice, KAVI = real-world broadcast speech.
}
\small
\begin{tabularx}{\textwidth}{p{0.20\textwidth} p{0.15\textwidth} p{0.10\textwidth} | XXX | XXX @{}}
\toprule
\multicolumn{1}{@{}l}{\multirow{2}{*}{\textbf{Backend}}} & \textbf{Config} & \multirow{2}{*}{\textbf{Params}} & \multicolumn{3}{c|}{\textbf{In-domain}} & \multicolumn{3}{c}{\textbf{Out-of-domain}} \\
\cmidrule(lr){4-6} \cmidrule(lr){7-9}
& \textbf{(L / Enc. / FFN)} & & \textbf{LP} & \textbf{FP'16} & \textbf{FP'20} & \textbf{CV} & \textbf{FLEURS} & \textbf{KAVI} \\
\midrule
\multicolumn{8}{@{}l}{\textbf{Base models}} \\
Transformer~\cite{getman25_interspeech} (baseline) & 12 / 768 / 3072 & 95.0M & 22.27 & 10.92 & 8.42 & 10.61 & 13.27 & 27.45 \\
Transformer (ours) & 12 / 768 / 3072 & 95.0M & 22.28 & 10.72 & 8.70 & 9.74 & 13.02 & 24.85 \\
 & 18 / 768 / 1536 & 95.1M & 21.33 & 10.64 & 8.66 & 8.99 & 12.71 & 23.91 \\
Conformer & 12 / 576 / 2048 & 96.7M & 18.89 & 10.08 & 8.07 & 6.46 & 11.20 & 23.71 \\
 & 17 / 512 / 1536 & 96.5M & \underline{18.50} & 9.83 & 7.94 & \underline{5.69} & \underline{10.56} & 22.71 \\
E-Branchformer & 12 / 576 / 1280 & 98.4M & 19.14 & 10.10 & \underline{7.80} & 6.91 & 11.53 & 22.89 \\
 & 17 / 512 / 1024 & 96.6M & 18.80 & \underline{9.54} & 7.85 & 6.19 & 11.09 & \underline{22.44} \\
\midrule
\multicolumn{8}{@{}l}{\textbf{Large models}} \\
Transformer~\cite{getman25_interspeech} (baseline) & 24 / 1024 / 4096 & 317.4M & 16.86 & 8.29 & 6.94 & 5.86 & 10.69 & 23.60 \\
Transformer (ours) & 24 / 1024 / 4096 & 317.4M & 16.78 & 8.41 & 6.77 & 6.20 & 10.42 & 21.02 \\
Conformer & 24 / 768 / 2608 & 317.2M & 16.46 & 8.58 & \underline{6.71} & 4.52 & 9.69 & \underline{19.86} \\
 & 36 / 640 / 2048 & 317.1M & 16.36 & 8.16 & 6.81 & \underline{4.28} & 9.43 & 21.02 \\
E-Branchformer & 24 / 768 / 1592 & 317.1M & 16.30 & 8.22 & 6.84 & 4.33 & 9.59 & 20.01 \\
 & 36 / 640 / 1280 & 317.4M & \underline{\textbf{16.11}} & \underline{8.15} & 6.88 & 5.24 & \underline{9.30} & 20.09 \\
\midrule
\multicolumn{8}{@{}l}{\textbf{X-Large models}} \\
Transformer~\cite{getman25_interspeech} (baseline) & 48 / 1280 / 4096 & 965.5M & 16.37 & 7.81 & 6.43 & 5.42 & 9.96 & 22.42 \\
Transformer (ours) & 48 / 1280 / 4096 & 965.5M & \underline{16.27} & 7.83 & \underline{\textbf{6.26}} & 3.85 & 9.37 & \underline{\textbf{19.54}} \\
Conformer & 48 / 1024 / 2768 & 964.7M & 17.13 & 7.78 & 6.45 & \underline{\textbf{3.68}} & \underline{\textbf{9.08}} & 20.75 \\
E-Branchformer & 48 / 1024 / 2048 & 965.6M & 16.74 & \underline{\textbf{7.63}} & 6.38 & 4.45 & 9.13 & 19.96 \\
\bottomrule
\end{tabularx}
\label{tab:architecture_exp_res_full}
\end{table*}

\subsection{Architecture comparison and depth-width trade-offs}

\label{subsec:arch_comparison}


Beyond data filtering, we analyze architectural modifications to improve parameter efficiency, evaluating alternative backends and depth-width configurations at fixed parameter budgets.

Table~\ref{tab:architecture_exp_res} presents the results for Base-sized models. Increasing depth generally improves performance at this scale. For Transformers, we observe a saturation point at 18 layers. Further increase to 24 layers with reduced width for parameter parity yields no improvement on 5 of 6 sets. 
Similar saturation patterns have been observed in language models~\cite{petty-etal-2024-impact}, suggesting that at fixed parameter budgets, the benefits of increased depth may be offset by the corresponding reduction in model width. 

Macaron-Net consistently outperforms Transformers. The 12-layer Macaron variant achieves lower WERs than the equivalent Transformer, while the 22-layer model further reduces error rates, outperforming all Transformer variants.


Convolution-augmented architectures achieve the largest improvements, particularly in out-of-domain generalization. Conformer yields the lowest error rates on cleaner read-speech benchmarks (CV: 5.69\%, FLEURS: 10.56\%), while E-Branchformer performs best on challenging broadcast data (KAVI: 22.44\%). 
As neither Conformer nor E-Branchformer consistently dominates, we opt to scale both to Large and X-Large sizes.
We additionally train Transformer models at all scales to enable direct comparison with prior work~\cite{getman25_interspeech}. 

\subsection{Scaling experiments}
\label{subsec:scaling_exp}


Table~\ref{tab:architecture_exp_res_full} presents results across all scales and benchmarks. 
Direct comparison with the prior Finnish wav2vec 2.0 baseline~\cite{getman25_interspeech} using identical Transformer configurations isolates the contribution of pretraining data. While the baseline used 158K hours primarily from clean talk radio with VAD segmentation, we retain 471K hours from 2.1M hours of raw, unconstrained broadcast archives through extensive confidence-based filtering. 

Our Transformer models achieve comparable in-domain performance to the baseline across all scales. Out-of-domain performance is generally improved, with most consistent gains on challenging broadcast data (KAVI), yielding 9.5-12.8\% relative WERR. These results show that our confidence-based filtering of unconstrained archives yields more effective pretraining than manually selecting cleaner sources with minimal preprocessing, improving generalization to real-world conditions without degrading in-domain performance.

While Base models benefit from deeper configurations at matched parameter count, this advantage does not persist at Large scale. The 36-layer variants achieve competitive or improved in-domain performance but yield inconsistent generalization: Conformer degrades 5.8\% relative on KAVI, and E-Branchformer underperforms 21\% relative on Common Voice compared to their 24-layer counterparts. This indicates that balancing model width with depth becomes critical for consistent stability and generalization at larger scales.


A complementary pattern is observed across encoder backends. At Base and Large scales, convolution-augmented encoders consistently outperform Transformer on out-of-domain benchmarks, with particularly large gains on Common Voice (31.0-36.7\% WERR). 
At X-Large scale, however, Transformer achieves the best KAVI result (19.54\%) and the most consistent scaling behavior, improving over Large across all benchmarks. In contrast, convolution-augmented encoders scale less uniformly, with domain-specific degradations (E-Branchformer on LP, Conformer on KAVI) relative to their Large counterparts. Overall, these findings indicate that architectural advantages provided by convolutional components are most pronounced in capacity-constrained regimes but attenuate as model scale approaches 1B parameters, 
where increased capacity narrows the performance gap between architectures.

\section{Conclusion}
\label{sec:conclusion}

We present a large-scale study of self-supervised speech representation learning for Finnish, retaining a high-quality 471K-hour subset from 2.1M hours of unconstrained broadcast archives. We show that raw broadcast material is a viable source for speech SSL when extensively filtered.
Our confidence-based filtering pipeline achieves 3-7\% WER improvements over LID-based filtering while naturally removing low-quality speech and foreign content.
For practitioners working with noisy multilingual broadcast data, our results support prioritizing strict filtering thresholds that favor quality over quantity.

Our systematic architectural analysis reveals that optimal design choices depend on parameter budgets. 
While deeper configurations and convolution-augmented encoders provide substantial improvements at Base scale (up to 37\% WERR), these advantages diminish as models approach 1B parameters, where standard Transformers scale most consistently.


Our optimized Base models substantially narrow the performance gap to larger variants relative to the prior Finnish SSL baselines, enabling effective ASR in resource-constrained settings. Out-of-domain performance continues improving through 1B parameters, suggesting larger models remain beneficial for diverse real-world conditions. 
While focused on Finnish, our findings provide guidance for SSL research on other languages with unconstrained audio sources.

\section{Acknowledgements}

The computational resources were provided by LUMI (project 462000971) and Aalto ScienceIT. We are grateful for the Business Finland project LAREINA under Grant 7817/31/2022.

\bibliographystyle{bibtex/bst/IEEEtran}
\bibliography{mybib}

\end{document}